\pdfoutput=1

\documentclass[11pt]{article}

\usepackage[]{acl}

\usepackage{times}
\usepackage{latexsym}

\usepackage[T1]{fontenc}

\usepackage[utf8]{inputenc}

\usepackage{microtype}

\usepackage{inconsolata}

\usepackage{graphicx}
\usepackage{multirow}
\usepackage{multicol}
\usepackage{subcaption}
\usepackage{enumitem,kantlipsum}
\usepackage[normalem]{ulem}
\usepackage{tabularx}
\usepackage{color, colortbl}
\usepackage{colortbl}
\usepackage{xcolor}
\usepackage{todonotes} 
\usepackage{rotating}
\usepackage{pbox}
\usepackage{booktabs} 
\usepackage{amssymb}
\usepackage{pifont}

\usepackage{tabu}
\usepackage{fontawesome}

\usepackage{amsmath}
\definecolor{shadowcolor}{rgb}{0.8, 0.8, 0.8}

\title{The FIGNEWS Shared Task on News Media Narratives}
  \author{Wajdi Zaghouani,$^1$ Mustafa Jarrar,$^2$ Nizar Habash,$^3$ Houda Bouamor,$^4$\\
\bf Imed Zitouni,$^5$ Mona Diab,$^6$ Samhaa R. El-Beltagy,$^7$ Muhammed AbuOdeh$^3$\\
$^1$Northwestern University in Qatar, Education City, Doha,
Qatar \quad 
$^2$Birzeit University, Palestine \\  
$^3$New York University Abu Dhabi, UAE \quad   
$^4$Carnegie Mellon University, Qatar \\   
$^5$Google, USA \quad  
$^6$Carnegie Mellon University, USA \quad   
$^7$Newgiza University, Egypt\\  
\texttt{wajdi.zaghouani@northwestern.edu, mjarrar@birzeit.edu, nizar.habash@nyu.edu} \\ \texttt{mdiab@cs.cmu.edu, samhaa@computer.org, mra9047@nyu.edu} \\
  }

\begin{document}
\maketitle
\begin{abstract}
We present an overview of the FIGNEWS shared task, organized as part of the {ArabicNLP~2024} conference co-located with ACL 2024.
The shared task addresses bias and propaganda annotation in multilingual news posts. We focus on the early days of the Israel War on Gaza as a case study.\footnote{FIGNEWS: \textit{Framing the Israel War on Gaza News}.} The task aims to foster collaboration in developing annotation guidelines for subjective tasks by creating frameworks for analyzing diverse narratives highlighting potential bias and propaganda. In a spirit of fostering and encouraging diversity, we address the problem from a multilingual perspective, namely within five languages: English, French, Arabic, Hebrew, and Hindi. A total of $17$ teams participated in two annotation subtasks: bias ($16$ teams) and propaganda ($6$ teams). The teams  competed in four evaluation tracks: guidelines development, annotation quality, annotation quantity, and consistency. Collectively, the teams produced $129,800$ data points. 
Key findings and implications for the field are discussed. 
\end{abstract}



\section{Introduction}
The FIGNEWS 2024 shared task,\footnote{Website: \url{https://sites.google.com/view/fignews}}$^,$\footnote{\label{foot-code-data}Code and Data: \url{https://github.com/CAMeL-Lab/FIGNEWS-2024} \&
\url{https://huggingface.co/datasets/CAMeL-Lab/FIGNEWS-2024}} henceforth FIGNEWS,
addresses the critical need for analyzing bias and propaganda in multilingual news discourse surrounding the Israel War on Gaza. This task aligns with the NLP community's growing efforts to create datasets and guidelines for complex opinion analysis through collaborative shared tasks and datathons. Such a meta-task allows for the exploration of various annotation frameworks in particular for complex and challenging subjective tasks. FIGNEWS focuses on a diverse multilingual corpus, emphasizing the development of guidelines with rich examples, while fostering a research-oriented collaborative environment.
By simultaneously examining multiple languages, comparing and contrasting various narratives, FIGNEWS aims to unravel the layers of possible bias and propaganda within news articles with alternative media narratives. This is especially critical at times of any war or conflict. The media's portrayal of events during such events has significant implications on public perception, policy-making, and international relations. By addressing bias and propaganda, we hope to illuminate the varied ways in which news can shape, and sometimes distort, public understanding of complex geopolitical events. This initiative seeks to explore diverse perspectives, cultures, and languages, thereby fostering a comprehensive understanding of events through the lens of major news outlets across the globe. 

Developing guidelines for complex data is a challenging task. The problem is exacerbated when the data is contemporaneous hence the annotators and the task organizers might have a stance on the subject matter. FIGNEWS is our attempt at addressing the creation of annotation guidelines, addressing what relevant best practices should be. We use the Israel War on Gaza as a use case to highlight some of these aspects. To that end, we curate a shared corpus for comprehensive annotation across various layers, crafting annotation guidelines shaped by the diverse range of conflicting discourses around this sensitive topic. This endeavor facilitates the development of robust methodologies and metrics for detecting and analyzing bias and propaganda, which are crucial for ensuring fair and accurate media reporting. The curated corpus, along with meticulously developed guidelines and annotations, will serve as a valuable resource for future research in NLP and related fields.

This initiative also seeks to bring to light both challenges and commendable aspects within the data, fostering a collaborative community that can learn from each other's approaches and findings. We believe that a collaborative, research-oriented environment is essential for tackling the intricate task of bias and propaganda detection. 

The FIGNEWS shared task thus represents a significant step forward in the field of data annotation, media analysis and NLP. A step towards an annotation science. It not only provides a platform for examining critical issues of bias and propaganda but also promotes the development of best practices in data annotation and analysis. Through this shared task, we aim to contribute to the broader goal of improving media literacy and fostering a more informed and critically engaged public.

\section{Related Work}

Propaganda and bias can have far-reaching implications depending on the context and medium in which they are propagated. Polarization, conflict and injustice are but some of the side effects they can create in any context. In the context of politics, they can alter the outcomes of elections and change the face of nations \cite{Gorenc2020, Maweu2019, Solopova2024}.    News media framing and narratives around wars and socio-political events have been extensively studied, with a focus on identifying bias, propaganda, offensive language detection, and diverse perspectives \cite{entman2007framing,baumer2015testing,fan2019plain,morstatter2018identifying,park2009newscube, DaSan2020, Aksenov2021,Yenkikar2022, Kameswari2020, Hong2023,Kim2023,Sharma2023, Maab2024,Rodrigo2024,HJKN23,DH21}. Several works have proposed computational approaches to detect media bias through analysis of word choice, labeling, and factual reporting \cite{hamborg2019automated,budak2016fair,vaagan2010tv,varacheva2018neutral}. 

\citet{entman2007framing} defines \textit{framing} as the process of selecting certain aspects of perceived reality and constructing a narrative that emphasizes their connections to promote a specific interpretation. This foundational work highlights how news framing can influence public perception by emphasizing particular elements over others. \citet{entman1993framing} further elaborates on this concept, providing a comprehensive framework for understanding how media frames can shape political and social realities.


Several studies have developed methods and datasets to detect news bias. \citet{budak2016fair} used crowdsourced content analysis to quantify media bias, while \citet{baumer2015testing} compared computational approaches for detecting framing in political news. Tools like \citet{Biasly2017} help users gauge news bias, quantifying liberal or conservative leanings.

In the context of multilingual and multi-label news framing analysis, \citet{akyurek2020multi} explored the complexities of news framing across different languages. Their work is crucial for understanding how bias and framing manifest in multilingual contexts, aligning closely with the objectives of the FIGNEWS shared task. Similarly, the study by \cite{Heppell2023} offers a valuable dataset and linguistic insights from two multilingual disinformation websites, providing a foundation for further exploration of language-specific disinformation techniques and their impact on news framing. 

Recent advances in bias detection have also emphasized the importance of detailed and well-annotated datasets. \citet{spinde2021towards} introduced the MBIC dataset, which includes detailed information about annotator characteristics and provides labels for bias identification at both the word and sentence levels. This dataset represents a significant step forward in creating reliable ground-truth data for bias detection. Additionally, \citet{spinde2021tassy} developed TASSY, a text annotation survey system that enhances the quality control of annotation processes in NLP tasks. 

Annotation of biased language and framing in news articles has been explored using techniques like expert annotation \cite{al-sarraj2018bias} and crowdsourcing \cite{lim2020annotating,lim2018understanding}. Quality control and guidelines for annotation processes in NLP tasks have also been investigated \cite{grosman2020eras,spinde2021tassy}. \citet{grosman2020eras} developed ERAS, a system designed to enhance quality control in NLP tasks, which is particularly relevant for ensuring the reliability of annotations in bias and propaganda detection.

Further contributions to the detection of bias and propaganda include the work of \citet{rashkin2017truth}, who analyzed language in fake news and political fact-checking, and \citet{barron2019proppy}, who organized news based on their propaganda-prone content. These studies provide foundational techniques for the identification and analysis of biased and propaganda language in news media.

The shared task on propaganda detection at SemEval-2020, organized by \citet{martino2020semeval}, is directly relevant to the FIGNEWS task. This task focused on detecting propaganda techniques in news articles, providing valuable benchmarks and methodologies that can be applied to the detection of bias and propaganda in social media posts.

In the context of the Israel-related wars and conflicts, \citet{al-sarraj2018bias} conducted a sentiment analysis study to detect bias in Western media coverage. This work highlights the specific challenges of detecting bias in a highly polarized and sensitive geopolitical context. Additionally, \citet{HJKN23} introduced a new dataset designed for detecting and classifying various forms of offensive language in Hebrew on social media.
Furthermore, the WojoodNER Shared Task 2024 offered a new NER dataset related to the Israeli War on Gaza called  \textit{Wojood\textsuperscript{Gaza}} \cite{JHKTEA24}.

%

Research on detecting political bias in social media includes \citet{zhou2020detecting}, who identified bias in user-generated content, and \citet{li2019survey}, who surveyed sentiment analysis techniques. The Propitter corpus \cite{Casavantes2024} was developed using distant supervision with refined annotations.
\citet{hamborg2019automated} advanced media bias detection by focusing on automated identification through word choice and labeling. \citet{fan2019plain} analyzed factual reporting to reduce bias, stressing the need for objectivity.

%
%
%

Among the studies forming a strong basis for understanding bias and propaganda detection in media are the following.
\citet{baly2020written} examined news media profiling using text and social media analysis, showing how news context affects bias perception. \citet{allcott2017social}, \citet{vosoughi2018spread}, and \citet{grinberg2019fake} studied fake news spread during the 2016 US election, highlighting the need for reliable information. 
\citet{lazer2018science} detailed fake news detection methods, while \citet{horne2017just} emphasized high-quality data for combating misinformation.
\citet{sharma2019combating} surveyed fake news identification techniques. \citet{zhou2020survey} reviewed fake news detection theories and methods, pointing to future research opportunities. 

While prior work has made notable contributions, our shared task focuses on comprehensive annotation of media narratives and framing around a specific war to uncover new insights. We build upon existing annotation frameworks \cite{zaghouani2016guidelines,zaghouani2018annotation} to develop improved guidelines and foster a collaborative exploration of media discourses through a multilingual, multicultural lens.

\section{Data Collection and Selection}
We used the CrowdTangle\footnote{\url{https://crowdtangle.com}} platform to collect Facebook posts related to the Israel War on Gaza in five languages: English, French, Arabic, Hebrew, and Hindi. Specifically, we retrieved public posts containing the keyword ``Gaza''\footnote{ \includegraphics[]{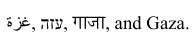}}
from verified blue-check accounts, publicly posted between October $7$, $2023$, and January $31$, $2024$.
%
%

For each language, we collected approximately $300,000$ posts. To narrow this down to $3,000$ posts per language and have a balanced distribution, we focused on ten key moments in the first few months of the war:
starting with the events of October $7$, $2023$ and the declaration of war, and including the bombings in Jabalia refugee camp, Al-Shifa hospital, and St. Porphyrius Church, as well as the ceasefire and hostage release, Gaza mass arrests, and the coverage of the International Court of Justice case  till the end of January $2024$.

For each of the ten moments, we selected the top $200$ posts per language, ranked by total interaction count (likes, shares, comments, etc.). As a result, we selected $15,000$ posts with the highest interactions, covering different key moments of the War, in five languages.

We divided the $15,000$ posts  into $15$ batches (\{\texttt{B01}, \texttt{B02}, ..., \texttt{B015}\}) with $1,000$ posts in each batch. Each batch contains $200$ posts per language. Additionally, each batch included $20$ posts for each of the $10$ key moments during the War.

Since the annotators may not speak all five languages, we provided machine translations (Google Translate) into English and Arabic to facilitate annotation across the multilingual corpus. While machine translation inevitably introduces some noise, we considered this to be reflective of the real-world reliance on such technologies.

To compare the Inter-Annotator Agreement (IAA) among all participating teams in a fair manner, we randomly selected 20 posts from each language (i.e.,  $100$ posts, which is $10$\%) from each batch, totaling $1,500$ posts. All participants received the dataset in a Google Sheet file with two sheets: ``Main'' and ``IAA''. The ``Main'' sheet contains $13,500$ posts, while the ``IAA'' sheet contains $1,500$ posts. See Appendix~\ref{sec:appendix:interface} for a screenshot of the interface.

It is important to note that participants were provided only with the original posts and their translations. Information such as account owner, date, and total interactions were not given to any participating team (See the ethical consideration section).

\begin{table*}[ht!]
\centering
\small

\begin{tabular}{|l | l | c |c|}
\hline
\textbf{Team Name} & \textbf{System Description} & \textbf{~~~~Bias~~~} & \textbf{Propaganda} \\
\hline
\textbf{Bias Bluff Busters} & \newcite{FIGNEWS:2024:BiasBluffBusters} & X & X \\
\hline
\textbf{BiasGanda} & \newcite{FIGNEWS:2024:BiasGanda} & X &  \\
\hline
\textbf{BSC-LANGTECH} & \newcite{FIGNEWS:2024:BSC-LANGTECH} & X &  \\
\hline
\textbf{Ceasefire} & \newcite{FIGNEWS:2024:Ceasefire} & X &  \\
\hline
\textbf{DRAGON} & \newcite{FIGNEWS:2024:DRAGON} & X &  \\
\hline
\textbf{Eagles} & \newcite{FIGNEWS:2024:Eagles} & X &  \\
\hline
\textbf{Groningen Annotates Gaza} & \newcite{FIGNEWS:2024:GroningenAnnotatesGaza} & X &  \\
\hline
\textbf{JusticeLeague} & \newcite{FIGNEWS:2024:JusticeLeague} & X &  \\
\hline
\textbf{Narrative Navigators} & \newcite{FIGNEWS:2024:NarrativeNavigators} & X & X \\
\hline
\textbf{NLPColab} & \newcite{FIGNEWS:2024:NLPColab} & X & X \\
\hline
\textbf{Sahara Pioneers} & \newcite{FIGNEWS:2024:SaharaPioneers} &  & X \\
\hline
\textbf{Sina} & \newcite{FIGNEWS:2024:Sina} & X & X \\
\hline
\textbf{SQUad} & \newcite{FIGNEWS:2024:SQUad} & X &  \\
\hline
\textbf{The CyberEquity Lab} & \newcite{FIGNEWS:2024:CyberEquityLab} & X & X \\
\hline
\textbf{The Guideline Specialists} & \newcite{FIGNEWS:2024:GuidelinesSpecialists} & X &  \\
\hline
\textbf{The Lexicon Ladies} & \newcite{FIGNEWS:2024:LexiconLadies} & X &  \\
\hline
\textbf{UoT1 }& \newcite{FIGNEWS:2024:UoT1} & X &  \\
\hline
\multicolumn{2}{r|}{\textit{Totals}}& 16 & 6 \\\cline{3-4}

\end{tabular}
 \caption{The participating teams: names, papers, and subtasks.}
    \label{tab:teams}
\end{table*}
%

\section{Subtasks and Evaluation Tracks}

This section presents the shared task subtasks, evaluation tracks and minimal requirement for teams to qualify. Further details are in Appendix~\ref{ST-details}.

\subsection{Minimal Requirements to Qualify}
To qualify, each participating team \textbf{must provide full annotation guidelines} for each subtask they choose to work on; and they \textbf{must  annotate at minimum Batch~1 and Batch~2}, i.e. (1,800 posts) and their designated Inter-annotator agreement subset (200 posts) for a total of 2,000 posts.

\subsection{Annotation Subtasks}

The shared task consists of two subtasks: Bias Annotation and Propaganda Annotation. 

\subsubsection{Bias Annotation Subtask}
The Bias Annotation subtask involves assigning one of the following seven labels to each post:
(1) \textbf{Unbiased},
(2) \textbf{Biased against Palestine},
(3) \textbf{Biased against Israel},
(4) \textbf{Biased against both Palestine and Israel},
(5) \textbf{Biased against others},
(6)  \textbf{Unclear}, and
(7) \textbf{Not Applicable}.
%
Examples illustrating each label are provided in the shared task description in Appendix~\ref{ST-details}.

\subsubsection{Propaganda Annotation Subtask}
The Propaganda Annotation subtask requires participants to classify each post into one of the following four categories:
(1) \textbf{Propaganda},
(2)  \textbf{Not Propaganda},
(3)  \textbf{Unclear}, and
(4) \textbf{Not Applicable}.
%
Examples showcasing each category are included in the shared task description in Appendix~\ref{ST-details}.

\subsection{Evaluation Tracks}
For each subtask, there are four evaluation tracks.

\subsubsection{Guidelines Track}
Participants have the freedom to design their own annotation guidelines and apply them to the shared data. The organizers will evaluate the guidelines based on an 8-point checklist, which includes items such as defining objectives, describing the task, establishing categories, providing detailed guidelines with examples, outlining the annotation process, setting quality standards, handling ambiguities, ensuring consistency, and considering ethical aspects.

The \textbf{Guidelines Score} used to determine the winners of this track is the average normalized Document Score and normalized IAA Kappa score.
The Document Score is equal to the number of satisfied document checklist items, a range from 0 to 8. The IAA Kappa score of a team is the average of all pairwise IAA Kappas over team annotators per batch (same as IAA Quality Score discussed next).  For both sub-scores, normalization is accomplished by dividing by the maximum value attained by any qualified team. Further details are in Appendix~\ref{ST-details}.

\subsubsection{IAA Quality Track}
\label{sec:iaa-quality}
In the IAA Quality  Track, teams compete based on their internal (within team) IAA Kappa scores \cite{Cohen:1960}. In addition to the Kappa score (our primary metric), we report on a number of other useful metrics. 

\begin{itemize}
\itemsep-3pt
\item \textbf{IAA Kappa}  (Primary Metric) A team's IAA Kappa score is calculated as the average of all pairwise Kappa scores between team annotators for each relevant IAA batch abd subtask.

    \item  \textbf{Accuracy (Acc)} The percentage of agreed upon data points (Bias and Propaganda)
   \item  \textbf{Macro F1 Average} The average F1 score over all the Bias or Propaganda subtask labels over all pairs of annotators and relevant IAA batches, i.e. batches which the pairs of annotators annotated.
   \item \textbf{F1 Bias*} The value of the average F1 score of all Bias labels collapsed as Bias vs other.
     \item \textbf{F1 Prop*} The value of the average F1 score of \textit{Propaganda} label alone. 
\end{itemize}

\subsubsection{Quantity Track}
In the Quantity Track, teams compete based on the number of annotated data points. They must complete the batches in order and finish one batch before moving to the next. 

\subsubsection{Consistency Track}
In the Consistency Track, teams compete based on the centrality of their annotation choices compared to all other teams. Centrality reflects the consistency of a team's annotations with those of other teams. We define a team's centrality as \textbf{Macro F1 Average} of its Bias or Propaganda annotations (as relevant) against other teams' annotations. A more central team, with higher consistency, is one that other teams agree with more on average.  We report on all the metrics mentioned in Section~\ref{sec:iaa-quality} except that we consider the annotations in Main Batch 1 and Batch 2 for this track, and only compare annotators in different teams (across teams).



\section{Results}
\subsection{Teams}
Out of the $23$ teams that registered, only $17$ technically qualified per the rules of the shared task, i.e., minimally provided batches~1 and ~2 in Main and IAA fully. Table~\ref{tab:teams} lists the qualifying teams and the subtasks they participated in. All of the qualifying teams submitted system description papers which are included in the proceedings.

\subsection{Annotators}
In total, 85 annotators from 16 teams participated in the Bias subtask, and 51 annotators from 6 teams participated in the Propaganda subtask.
Table~\ref{fig:table_annotators} 
in Appendix~\ref{sec:appendix:anno-demo} 
highlights the diversity among the annotators based on the information they provided voluntarily.  Only half of the annotators are native Arabic speakers and close to one-third are Urdu speakers. Half are between the ages 18-24 and close to one-third 25-34. Over three-quarters identify as female. They claim many regions of origin (South Asia 33\%, Levant 27\%, North Africa 13\%, Western Europe 11\%, among others).  Almost all are highly educate with close to half with Master's degree. Two thirds of the annotators come from Engineering and Technology areas of expertise.

\begin{table*}[ht]
\centering
 \includegraphics[width=1\textwidth]{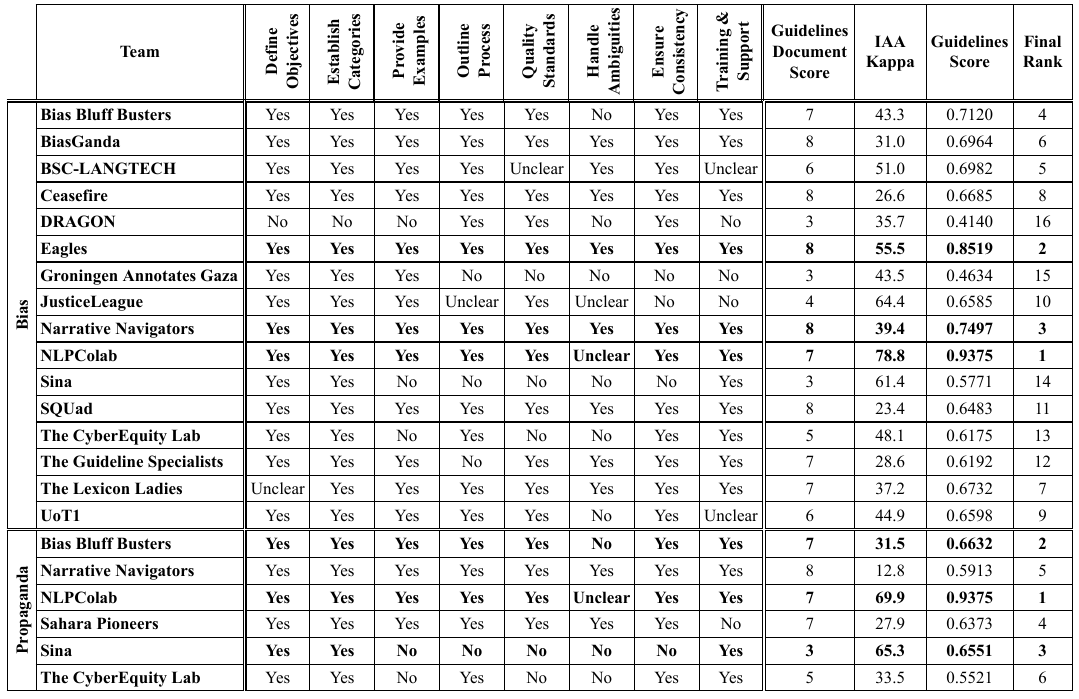}
   \caption{Results of the guidelines evaluation track}
\label{fig:table_guidelines_results}
\end{table*}

\begin{table*}[t]
\centering
 \includegraphics[width=1\textwidth]{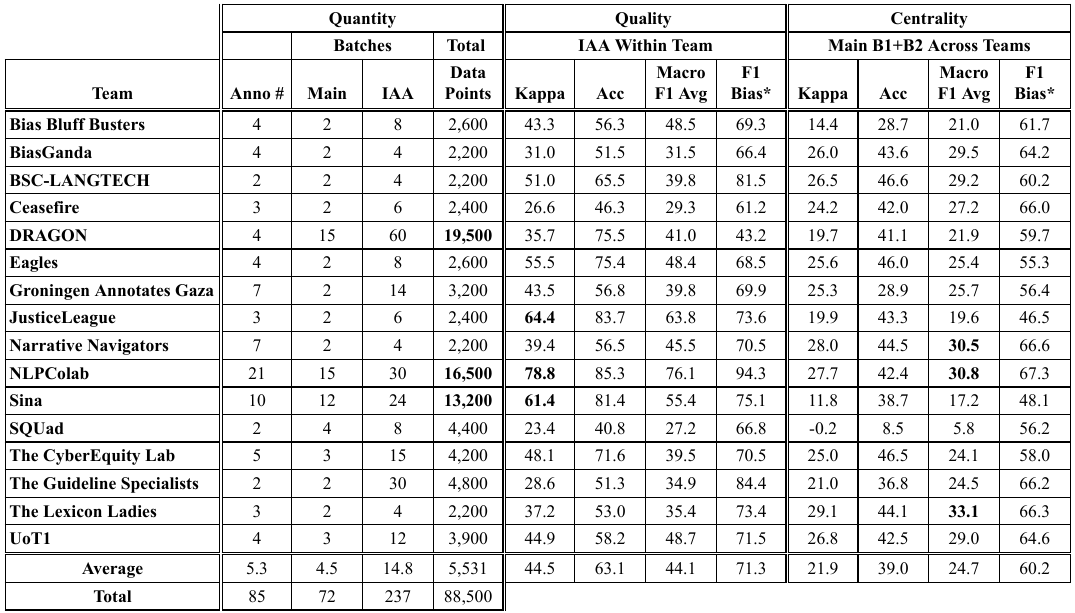}
    \caption{Results of the Bias subtask.}
\label{tab:bias-results}
\end{table*}
\begin{table*}[h!t]
\centering
 \includegraphics[width=0.95\textwidth]{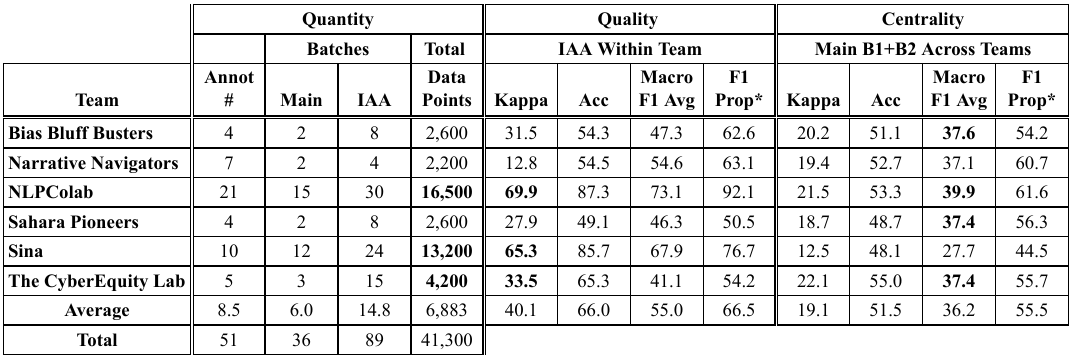}
    \caption{Results of the Propaganda subtask.}
\label{tab:propaganda-results}
\end{table*}

\subsection{Bias Subtask}

In total, there were $85$ annotators across $16$ teams and they annotated together $72$ Main sets and $237$ IAA sets, for a total of $88,500$ data points.
Table~\ref{fig:table_guidelines_results} and Table~\ref{tab:bias-results} present the results on the Bias Subtask. The winners are presented in Table~\ref{fig:table_results_summary}.

\paragraph{Guidelines Track}
The winners of the Bias Guidelines Track are NLPColab (1st), Eagles (2nd) and Narrative Navigators (3rd).  Details on their scores and ranking are in Table~\ref{fig:table_guidelines_results}.

\paragraph{IAA Quality Track}
The winners of the Bias IAA Quality Track are NLPColab (1st), JusticeLeague (2nd) and Sina (3rd).  Details on their scores are in 
Table~\ref{tab:bias-results}.

\paragraph{Quantity Track} 
The winners of the Bias Quantity Track are DRAGON (1st), NLPColab (2nd) and Sina (3rd). The sum of their annotations equal to 56\% of all Bias subtask annotations.  Details on their scores are in 
Table~\ref{tab:bias-results}.

\paragraph{Consistency Track} 
The winners of the Bias Consistency Track are The Lexicon Ladies (1st), NLPColab (2nd) and Narrative Navigators (3rd). Details on their scores are in 
Table~\ref{tab:bias-results}.

\paragraph{Observations} 
As anticipated, within-team IAA scores significantly surpass across-team IAA, with an average absolute increase of 22.6\% (\textbf{Kappa}) and 19.4\% (\textbf{Macro F1 Average}). The \textbf{F1 Bias*} scores, indicating binary bias determination, show a wide range of disagreements within teams (average 71.3\%, stdev 10.9\%) and across teams (average 60.2\%, stdev 6.5\%). This variability reflects the inherent subjectivity and complexity of bias labeling in general.

The bias labeling task is challenging, with high IAA difficult to achieve both within and across teams. However, the high scores of top-performing teams highlight the need for meticulous attention to detail and comprehensive training.

\subsection{Propaganda Subtask}

In total, there were $51$ annotators across $6$ teams and they annotated together $36$ Main sets and $89$ IAA sets, for a total of $41,300$ data points.
Table~\ref{fig:table_guidelines_results} and Table~\ref{tab:propaganda-results} present the results on the Propaganda Subtask. The winners are presented in Table~\ref{fig:table_results_summary}.

\paragraph{Guidelines Track}
The winners of the Propaganda Guidelines Track are NLPColab (1st), Bias Bluff Busters (2nd) and Sina (3rd).  Details on their scores and ranking are in Table~\ref{fig:table_guidelines_results}.

\paragraph{IAA Quality Track}
The winners of the Propaganda IAA Quality Track are NLPColab (1st), Sina (2nd) and The CyberEquity Lab (3rd).  Details on their scores are in 
Table~\ref{tab:propaganda-results}.

\paragraph{Quantity Track} 
The winners of the Propaganda Quantity Track are NLPColab (1st), Sina (2nd) and The CyberEquity Lab (3rd). The sum of their annotations equal to 82.1\% of all Propaganda subtask annotations.  Details on their scores are in 
Table~\ref{tab:propaganda-results}.

\paragraph{Consistency Track} 
The winners of the Propaganda Consistency Track are NLPColab (1st), Bias Bluff Busters (2nd) and Sahara Pioneers and The CyberEquity Lab (tied 3rd). Details on their scores are in 
Table~\ref{tab:propaganda-results}.

\paragraph{Observations} 
As anticipated, within-team IAA scores significantly surpass across-team IAA, with an average absolute increase of over 21.1\% (\textbf{Kappa}) and 18.9\% (\textbf{Macro F1 Average}). The \textbf{F1 Prop*} scores show a wide range of disagreements within teams (average 66.5\%, stdev 15.4\%) and across teams (average 55.5\%, stdev 6.1\%). This variability reflects the inherent subjectivity and complexity of this task.

Like Bias labeling, the Propaganda labeling task is quite demanding. Although it has a smaller number of labels, we see comparable patterns in terms of performance across a number of metrics.


\section{Discussion}

\begin{table*}[t]
\centering
 \includegraphics[width=0.85\textwidth]{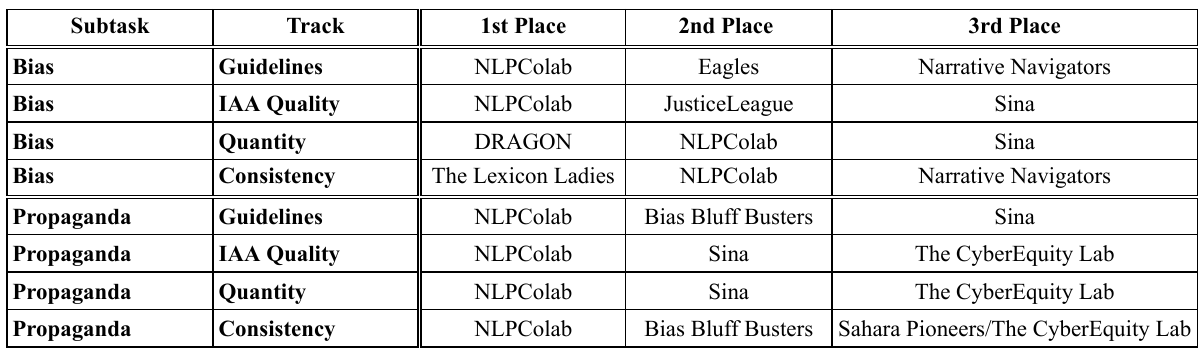}
    \caption{Shared task winners for each subtask and track.}
\label{fig:table_results_summary}
\end{table*}
\begin{table*}[t]
\centering
 \includegraphics[width=0.85\textwidth]{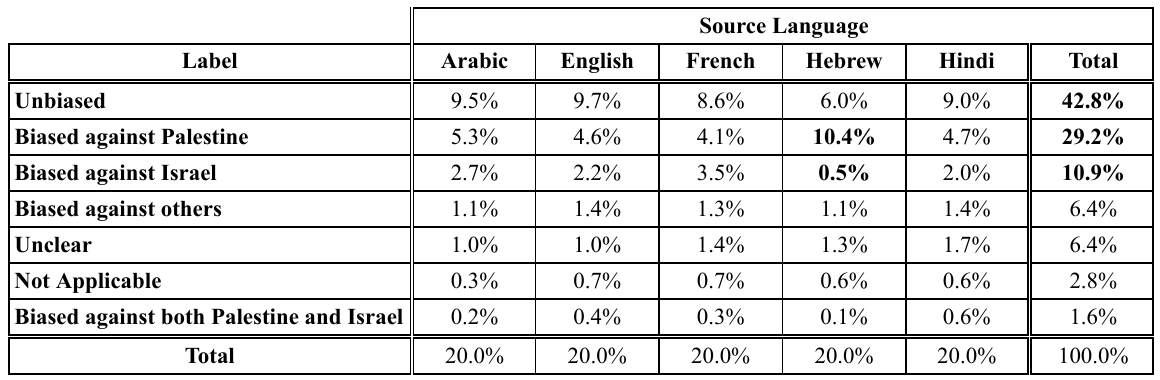}
    \caption{Bias Label Distributions in total and over source language.}
\label{tab:bias-patterns}
\end{table*}
\begin{table*}[th!]
\centering
 \includegraphics[width=0.75\textwidth]{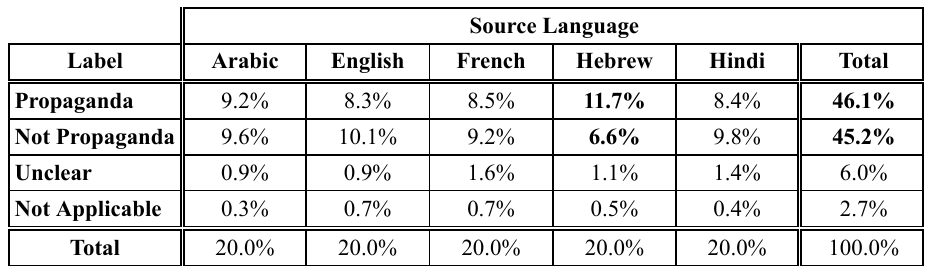}
    \caption{Propaganda Label Distributions in total and over source language.}
\label{tab:prop-patterns}
\end{table*}
\begin{table*}[th!]
\centering
 \includegraphics[width=0.85\textwidth]{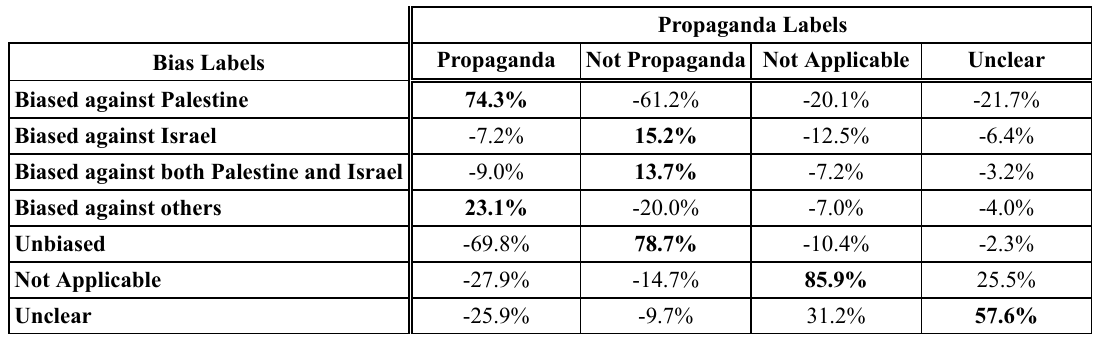}
    \caption{Pearson correlation coefficient for all Bias vs Propaganda label pairs. }
\label{tab:bias-propaganda-patterns}
\end{table*}

\begin{table*}[th]
\centering
  \includegraphics[width=0.92\textwidth]{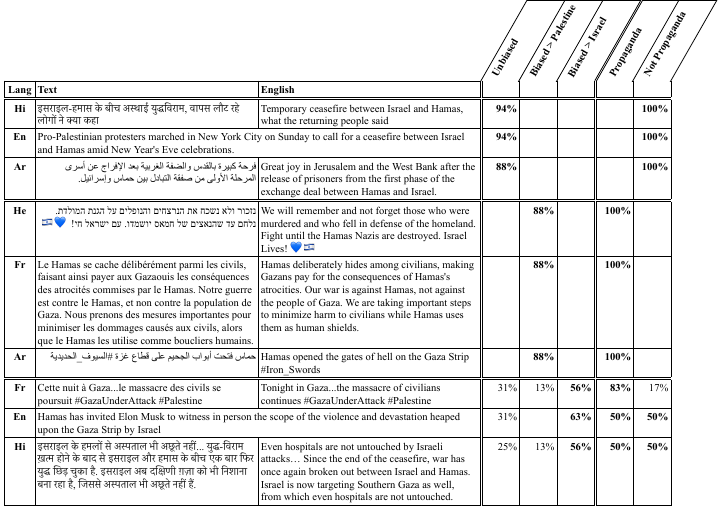}
    \caption{Examples of different texts and their most relevant annotation distributions.}
\label{fig:example}
\end{table*}

We now consider the label distribution patterns for Bias and Propaganda, independently and together.

\subsection{Bias Label Distributions}
Table~\ref{tab:bias-patterns}  summarizes the Bias label distributions overall and for different source languages: Arabic (Ar), English (En), French (Fr), Hebrew (He), and Hindi (Hi).
The reported results here include all annotated data points (from Main and IAA). Naturally,  data points from Batches 1 and 2 are over-represented  since they were annotated by all teams. 
\textbf{Unbiased} was the highest overall label claiming over two-fifths of all data points.
This is followed by \textbf{Biased against Palestine} (29.2\%), then \textbf{Biased against Israel}; with 
\textbf{Biased against Palestine} almost three times that of \textbf{Biased against Israel}.
\textbf{Biased against others} appeared in about 1/16th of the cases.
All labels seem to be generally equally distributed across source languages with the exception of \textbf{Biased against Palestine} and \textbf{Biased against Israel}  standing out in Hebrew-sourced texts: \textbf{Biased against Palestine} is twice the average of other languages; and \textbf{Biased against Israel} is one-fifth the average of other languages.
While this seems consistent with what would be expected of the news media of a country at war, it is possible that there is an additional priming bias from knowing the source language of the text. Perhaps in future editions, we could compare with a setting where all source language information is hidden and only translations are provided.

\subsection{Propaganda Label Distributions}
Table~\ref{tab:prop-patterns}  summarizes the Propaganda label distributions overall and for different source languages. The reported results here include all annotated data points (from Main and IAA).
\textbf{Propaganda} and \textbf{Not Propaganda} split the distribution almost equally with \textbf{Propaganda} being slightly higher. 
All labels seem to be generally equally distributed with the exception of of Hebrew-sourced texts where propaganda spikes:
the ratio of \textbf{Propaganda} to  \textbf{Not Propaganda} is 1.8 as opposed to less than 0.9 for the other source languages.  The same observation made in the previous section applies here.

\subsection{Bias vs Propaganda Label Correlations}    
Finally, Table~\ref{tab:bias-propaganda-patterns}
presents a cross-comparison of Bias and Propaganda labels using Main Batches 1 and 2 ($1,800$ data points).
For each text in this data subset, we calculate for each label in Bias and Propaganda subtasks the number of teams that selected that label. We then report the Pearson product-moment correlation coefficient of the $1,800$ counts for every pair of Bias-Propaganda labels.
Unsurprisingly, Bias \textbf{Not Applicable} and Propaganda \textbf{Not Applicable}  relate closely (r=85.9\%).  Similarly r=78.7\%  for \textbf{Unbiased} is \textbf{Not Propaganda}.
The \textbf{Biased against Palestine} label correlates positively highly (74.3\%) with \textbf{Propaganda}, and negatively strongly (-61.2\%) with \textbf{Not Propaganda}. The patterns are reversed and much weaker for
\textbf{Biased against Israel}.  These are only high level initial observations that open exciting possibilities for further study.

\subsection{Selected Examples}
Table~\ref{fig:example} presents a number of texts with their associated most prominent label averages across the participating teams. The examples are from Main Batches 1 and 2.  There are nine examples; the first three were marked by the vast majority of teams as \textbf{Unbiased} and \textbf{Not Propaganda}. The second set of three examples were marked by the vast  majority again as \textbf{Biased against Palestine} and \textbf{Propaganda}. The last set shows examples with majority \textbf{Biased against Israel}.


    


\section{Conclusion and Future Work}
The FIGNEWS shared task successfully brought together a diverse community to annotate bias and propaganda in multilingual news posts. This initiative brought together 17 teams producing 129,800 data points.  The shared task highlighted the crucial role of clear guidelines, examples, and collaboration in advancing NLP research on complex, subjective, and sensitive, opinion analysis tasks. The resulting dataset and insights contribute valuable resources and direction for future work in this important area. All data and code are publicly available.\textsuperscript{\ref{foot-code-data}}
 
Future work should focus on expanding the annotation efforts to include more diverse languages and topics, and refining annotation guidelines based on participant feedback. The created data 
should be leveraged to  advance NLP automatic bias and propaganda detection techniques, as well as foster interdisciplinary studies  to deepen our understanding of bias and propaganda in news media.

\section*{Limitations}
We acknowledge the following limitations in the FIGNEWS shared task design and implementation.
\begin{itemize} 
\itemsep-3pt
    \item  \textbf{Annotation Subjectivity} The task involves subjective judgments on bias and propaganda, which vary among annotators and teams, impacting annotation consistency and reliability. Teams' self-selection based on preconceived notions about the topic may further influence this variability.
    \item \textbf{Label Selection} We acknowledge that the set of labels we specified limit the space of possibilities and may oversimplify complex issues imposing a binary perspective that does not fully capture nuanced viewpoints and biases within the dataset.
    \item \textbf{Scope of Topics and Text Selection} The focus on the early days of the Israel War on Gaza may limit the applicability of findings to other types of news events or broader media contexts. The size of the corpus is relatively small, and may include some sampling bias.
    \item \textbf{Limited Diversity} while we observed a range of backgrounds among the annotators, we acknowledge that some groups were highly over-represented, which potentially biases the overall conclusions.
    
\end{itemize}

\section*{Ethical Considerations}
\label{sec_ethics}
The FIGNEWS shared task deals with sensitive topics and media narratives related to the Israel War on Gaza. The organizers and participants have taken several measures to ensure ethical considerations are addressed:

\begin{itemize}
\itemsep-3pt
\item \textbf{Anonymization} All posts have been anonymized, with no identifying information about the account owners or users provided to the participants.

\item \textbf{Public Posts} Only publicly available posts from verified accounts were included in the dataset, ensuring that the content was intended for public dissemination.

\item \textbf{Balanced Representation} To ensure fair representation, the dataset includes a balanced number of posts from various viewpoints and narratives during the war.

\item \textbf{Responsible Use} Participants were required to agree to use the dataset solely for research purposes and not for any unethical or illegal activities.
\end{itemize}

\section*{Acknowledgments}
This task is partially supported by NPRP 14C-0916-210015 from the Qatar National Research Fund, which is a part of Qatar Research Development and Innovation Council (QRDI). The findings achieved herein are solely the responsibility of the authors.

\bibliography{bib/references}
\bibstyle{acl_natbib}

\onecolumn
\appendix

\section{Shared Task Details}
\label{ST-details}
We provide below an updated version of the shared task details reflecting the final decisions made in the effort, e.g., we added a fourth evaluation track (IAA Quality) that was not mentioned in the original call to participate.

\subsection{Shared Task Objectives}
The shared task aims to serve as a collaborative platform where participants propose guidelines and diverse methods for annotating and analyzing the dataset. 

\paragraph{Provided Data}
The organizers will provide 15 batches of social media posts, 1,000 post per batch.
Each 1,000-post batch will contain 200 posts from 5 languages: Arabic, English, French, Hebrew, and Hindi, together with their machine translated versions into Arabic and English (as needed).  Participants must specify whether they annotated the original language or its machine translated version. So a monolingual Arabic team can annotate the full batch in Arabic (original or translated).
The batches will be provided to the annotation teams with clear instruction on how to submit the results.

\paragraph{Minimal Annotations to Qualify}
To qualify, each participating team \textbf{must provide full annotation guidelines} for each subtask they choose to work on; and they \textbf{must  annotate  a minimum of two batches}, i.e. (1,800 posts) and their designated Inter-annotator agreement subset (200 posts) for a total of 2,000 posts (specifically Batch~1 and Batch~2). The IAA subset must be done by every annotator on the team; but the rest can be divided among them.

\subsection{Shared Task Subtasks and Tracks}
There will be two \textbf{subtasks} of focus.
For each subtask, there will be four \textbf{evaluation tracks} for which winners will be crowned.

\subsubsection{Subtask on Bias Annotation}

The subtask is restricted to seven  possible Labels, presented below with illustrative examples.

\begin{enumerate}
\itemsep-2pt
\item \textbf{Unbiased}\\
Example: "In the ongoing Israel-Palestine conflict, recent events have escalated tensions. Yesterday, Israeli forces conducted operations in response to rocket attacks from Gaza. Both sides have reported casualties. International leaders are calling for restraint and a return to peace talks."

\item \textbf{Biased against Palestine}\\
Example: "Once again, Palestinian aggression has disrupted peace in the region. Palestinian extremists, ignoring efforts for peace, launched unprovoked attacks on innocent Israeli civilians. Israel's response, though portrayed as harsh by some, is a justified measure to protect its citizens."

\item \textbf{Biased against Israel}\\
Example: "In a typical display of excessive force, Israeli troops have yet again targeted Palestinian areas, causing numerous civilian casualties. This aggression, under the guise of self-defense, highlights the ongoing oppressive tactics Israel employs against Palestinians."

\item \textbf{Biased against both Palestine and Israel}\\
Example: "In the latest chapter of their endless and futile conflict, Israeli and Palestinian forces have once again engaged in senseless violence. Both sides continue to commit atrocities, showing a complete disregard for peace or human life."

\item \textbf{Biased against others}\\
Example: "In the shadow of the Israel-Palestine conflict, external actors, particularly Iran, are exacerbating tensions. Iran's covert support for extremist groups shows its intent to destabilize the region, disregarding the catastrophic impact on both Israeli and Palestinian civilians."

\item \textbf{Unclear}\\
Example: "Recent developments in the Middle East have seen an increase in hostilities. The situation in the region is complex, with various factors contributing to the current state of affairs. The international community remains divided on the issue."
 
\item \textbf{Not Applicable}\\
Example: "In other news, the annual technology conference in Tel Aviv has unveiled groundbreaking advancements in cybersecurity. Industry leaders from around the globe gathered to showcase innovations that promise to shape the future of digital security."
\end{enumerate}

\subsubsection{Subtask on Propaganda Annotation}

The subtask is restricted to four possible Labels, presented below with illustrative examples.

\begin{enumerate}
\itemsep-2pt
\item \textbf{Propaganda}\\
Example: "In a display of unmatched heroism, our troops have once again safeguarded our nation from the brink of destruction, heroically neutralizing the threat from Gaza, which aims to undermine our very existence."
\item \textbf{Not Propaganda}\\
Example: "Yesterday, an escalation occurred along the Israel-Gaza border, resulting in casualties on both sides. Israeli and Palestinian officials provided conflicting accounts of the events that led to the confrontation."
\item \textbf{Unclear}\\
Example: "The situation in Gaza remains tense, with reports of civilian distress and military movements. While some sources claim the military actions are defensive, others argue they are provocative, leaving the true nature of the situation open to interpretation."
\item \textbf{Not Applicable}\\
Example: "A feature on Gaza's cultural scene highlights the resilience of its art community, showcasing how local artists use their craft to express hope and endurance amid challenging circumstances, without delving into the political context."
\end{enumerate}

\subsubsection{Guidelines Evaluation Track}
The teams have the freedom to design their own guidelines and apply them to the shared data. The following is the checklist of all items that will be evaluated by the organizers.

\paragraph{Annotation Guidelines}
Detailed annotation guidelines including examples for all main and corner cases.
Consider the following components which will be used in the evaluation of the guidelines.

\begin{enumerate}
\itemsep-2pt
\item \textbf{Define the Objective and Describe the Task}\\
Outline the purpose and specific NLP task. Provide a detailed task description with correct examples.

\item \textbf{Establish Categories}\\
List and define all annotation labels/categories/tags.

\item \textbf{Include Detailed Category Guidelines with Examples}\\
Explain application criteria for each category/tag, with examples.
Offer examples for correct application and common mistakes.

\item \textbf{Outline the Process}\\
Describe the step-by-step annotation process and tools used.

\item \textbf{Set Quality Standards}\\
Define expectations for accuracy and consistency, along with quality check procedures.

\item \textbf{Handle Ambiguities and Difficult Cases}\\
Provide guidance on ambiguous cases and a protocol for seeking clarification.

\item \textbf{Ensure Consistency}\\
Implement measures for annotator consistency and recommend calibration sessions.

\item \textbf{Training and Support}\\
Detail training procedures and support resources for annotators.
Highlight unbiased annotation practices and handling of sensitive data.
Schedule guideline reviews for updates based on feedback and new insights.
Include a system for annotator feedback to refine guidelines and processes.
\end{enumerate}

 The teams must provide well-documented annotation guidelines including examples, and must provide inter-annotator agreement (IAA) numbers for at least 200 posts (40 from each language) from Batch 1 and Batch 2.  We expect the IAA to be competitive (e.g. Cohen Kappa of 0.6+) in the target space. 
 %
 
The Guidelines Score used to determine the winners of this track is the average normalized Document Score  and IAA Kappa score.

\[
\text{GuidelinesScore$_i$} = \text{Average}\left( \frac{\text{DocumentScore$_i$}}{\text{DocumentScore$_{max}$}}, \frac{\text{IAAKappa$_i$}}{\text{IAAKappa$_{max}$}} \right)
\]

The Document Score is equal to the number of satisfied document components mentioned above, a range from 0 to 8. The IAA Kappa score of a team is the average of all pairwise IAA Kappas over team annotators per batch.

\subsubsection{IAA Quality Evaluation Track}
We will also report the IAA Kappa scores per team and use them to determine the best performers, independent of the guidelines track.

\subsubsection{Quantity Evaluation Track}
The teams can compete in the number of annotated data batches. They must finish them in order and complete a batch before moving to the next.  The teams with the highest number of completed batches will be crowned the Quantity track winners for the subtask of choice. 

\subsubsection{Consistency Evaluation Track}
The various teams in the same subtask and shared completed batches will be compared for correlation against each other. The teams that have the highest correlation against other teams (centroidal choices) will be crowned winner. This needs a minimum of three teams per subtask.

\subsection{Publication} All teams participating in the shared task are invited to submit short paper (4 pages) descriptions of their efforts. These papers will be evaluated by multiple reviewers to be selected for publication in the ArabicNLP 2024 Conference Proceedings and indexed by the ACL Anthology.

\subsection{Collaborative Commitment}
Participants are encouraged to join the shared task with a commitment to collaboration. Whether working independently or within teams, every effort and insight contributed should be shared openly. This collaborative ethos extends beyond individual tasks and includes sharing methodologies, findings, and results.

\subsection{Optional Demographic Details} We would like to invite participants to provide some demographic details voluntarily. This information includes aspects such as age range, native language, educational background, area of study or expertise, gender, and region of origin. Please note that providing this demographic information is entirely optional and will not influence the evaluation of your participation in any way. We respect your privacy and understand if you choose not to share these details.

\newpage
\section{Annotation Interface}
\label{sec:appendix:interface}

\begin{figure*}[ht]
\centering
 \includegraphics[width=1\textwidth]{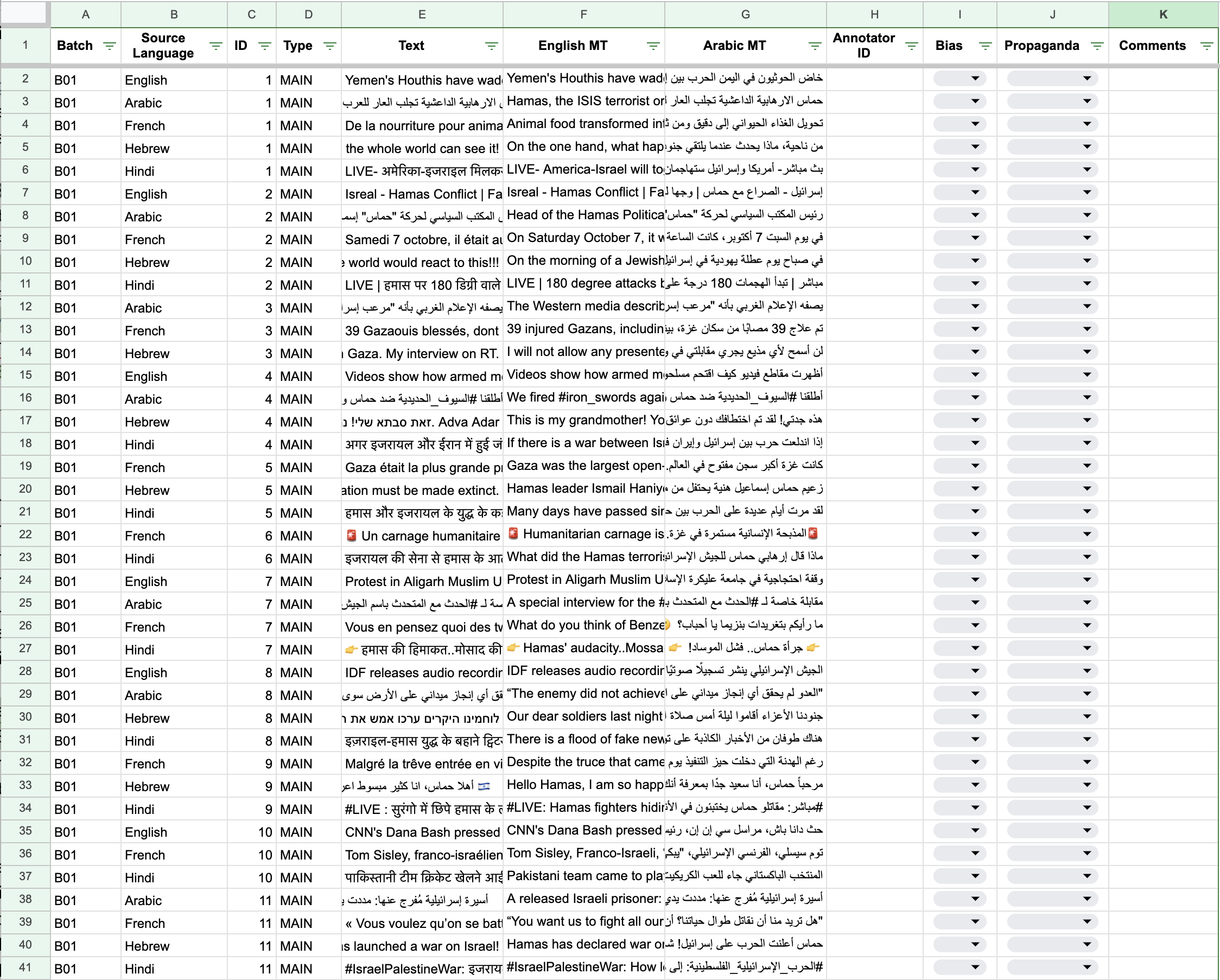}
    \caption{A screenshot of the Google Sheet annotation setup for the Main data subset.}
\label{fig:interface}
\end{figure*}

\newpage
\section{Annotator Demographics}
\label{sec:appendix:anno-demo}
\begin{table*}[ht]
\centering
 \includegraphics[width=1\textwidth]{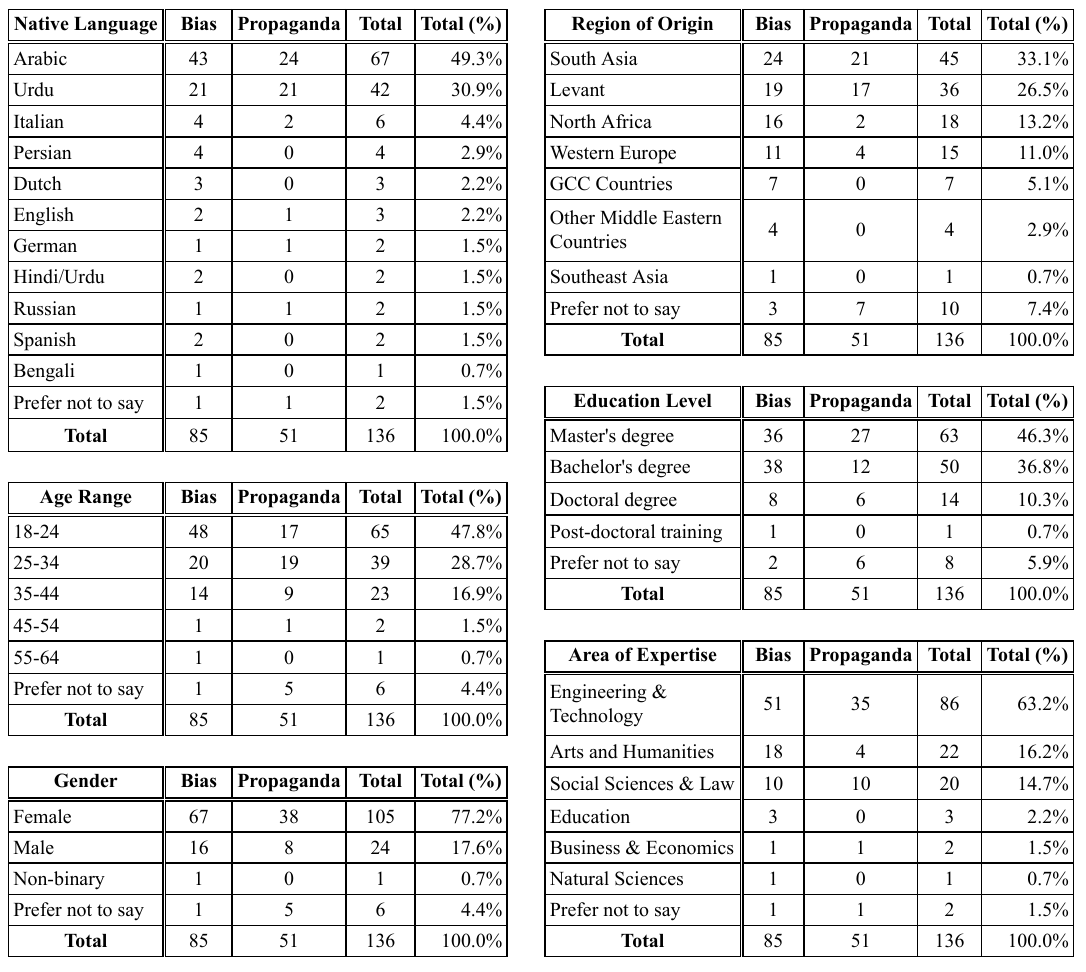}
    \caption{Annotator demographics.}
\label{fig:table_annotators}
\end{table*}







\end{document}